\documentclass[runningheads]{llncs}
\usepackage{graphicx}
\usepackage{tikz}
\usepackage{comment}
\usepackage{amsmath,amssymb} % define this before the line numbering.
\usepackage{color}

\usepackage[colorlinks=true, allcolors=blue]{hyperref}
\usepackage{url}

\newcommand{\ie}{i.e.}
\newcommand{\eg}{e.g.}

\usepackage{epsfig}
\usepackage{graphicx}

\usepackage{booktabs}
\usepackage{multirow}
\usepackage{subfig}
\usepackage{wrapfig}

\usepackage[accsupp]{axessibility}  % Improves PDF readability for those with disabilities.

\begin{document}
% \linenumbers
\pagestyle{headings}
\mainmatter
\def\ECCVSubNumber{776}  % Insert your submission number here

%\title{An Empirical Study of Distilling\\ Unconditional GANs} % Replace with your title
\title{Mind the Gap in Distilling StyleGANs} % Replace with your title

% %\begin{comment}
% \titlerunning{ECCV-22 submission ID \ECCVSubNumber} 
% \authorrunning{ECCV-22 submission ID \ECCVSubNumber} 
% \author{Anonymous ECCV submission}
% \institute{Paper ID \ECCVSubNumber}
% %\end{comment}
%******************

% \author{Guodong Xu\inst{1}\orcidID{0000-0002-7026-1375} \and
% Yuenan Hou\inst{1}\orcidID{0000-0002-2844-7416} \and
% Ziwei Liu\inst{2}\orcidID{0000-0002-4220-5958} \and \\
% Chen Change Loy\inst{2}\orcidID{0000-0001-5345-1591}}

% \begin{comment}
\author{Guodong Xu\inst{1} \and
Yuenan Hou\inst{2} \and
Ziwei Liu\inst{3} \and 
Chen Change Loy\inst{3}}
\authorrunning{G. Xu et al.}
% \institute{The Chinese University of Hong Kong \\
% \email{xg018@ie.cuhk.edu.hk} \and
% Shanghai AI Lab \\
% \email{houyuenan@pjlab.org.cn}\and
% Nanyang Technological University \\
% \email{\{ziwei.liu, ccloy\}@ntu.edu.sg}}

\institute{The Chinese University of Hong Kong \and Shanghai AI Laboratory \\
\and S-Lab, Nanyang Technological University \\
\email{xg018@ie.cuhk.edu.hk, houyuenan@pjlab.org.cn, \\
\{ziwei.liu, ccloy\}@ntu.edu.sg}}

% \end{comment}
%******************
\maketitle

% !TEX root = ../iclr2022_conference.tex

\begin{abstract}

StyleGAN family is one of the most popular Generative Adversarial Networks (GANs) for unconditional generation. Despite its impressive performance, its high demand on storage and computation impedes their deployment on resource-constrained devices.
This paper provides a comprehensive study of distilling from the popular StyleGAN-like architecture.
Our key insight is that the main challenge of StyleGAN distillation lies in the output discrepancy issue, where the teacher and student model yield different outputs given the same input latent code.
Standard knowledge distillation losses typically fail under this heterogeneous distillation scenario.
We conduct thorough analysis about the reasons and effects of this discrepancy issue, and identify that the mapping network plays a vital role in determining semantic information of generated images.
Based on this finding, we propose a novel initialization strategy for the student model, which can ensure the output consistency to the maximum extent.
To further enhance the semantic consistency between the teacher and student model, we present a latent-direction-based distillation loss that preserves the semantic relations in latent space.
Extensive experiments demonstrate the effectiveness of our approach in distilling StyleGAN2 and StyleGAN3, outperforming existing GAN distillation methods by a large margin. Code is available at: \href{https://github.com/xuguodong03/StyleKD}{https://github.com/xuguodong03/StyleKD}

\end{abstract}

% !TEX root = ../iclr2022_conference.tex

\section{Introduction}
\label{sec:intro}

GAN compression~\cite{ganslimming,gan_compression,CAGAN} has been actively studied to enable the practical deployment of powerful GAN models~\cite{pggan,stylegan,stylegan2} on mobile applications and edge devices.
%Generative adversarial networks (GANs)~\cite{GAN} have achieved state-of-the art results in many computer vision tasks at the expense of a drastic increase in network complexity and computation.
%
%To facilitate the deployment of powerful GANs in resource-limited scenarios, GAN compression~\cite{ganslimming,gan_compression,CAGAN} has become an active research topic in recent years. 
%
Among these techniques, knowledge distillation (KD)~\cite{KD} is a widely adopted training strategy for GAN compression.
The objective of GAN distillation is to transfer the rich dark knowledge from the original model (teacher) to the compressed model (student) so as to mitigate the performance gap between these two models.
There are two distillation strategies, \ie, pixel-level and distribution-level. The former minimizes the distance between generated images of two models, while the latter minimizes the distance between distributions. In this work, we focus on the first setting considering its prevalence in the GAN compression literature~\cite{portable,ganslimming,gan_compression,CAGAN}.

The majority of contemporary GAN distillation methods~\cite{portable,ganslimming,gan_compression,CAGAN} focus on conditional GANs (cGANs), especially image-to-image translation~\cite{pix2pix,cyclegan}, while the distillation of unconditional GANs (uncGANs) is relatively under-explored. 
Since there is a large difference between the learning dynamics of these two types of GANs, distillation methods tailored for cGANs cannot be directly applied to the unconditional setting.

\begin{figure}[t]
    \begin{minipage}[b]{.33\linewidth}
        \centering
        \subfloat[][Dog-Cat classification]{\label{fig:dog_cat}
        \includegraphics[scale=0.25]{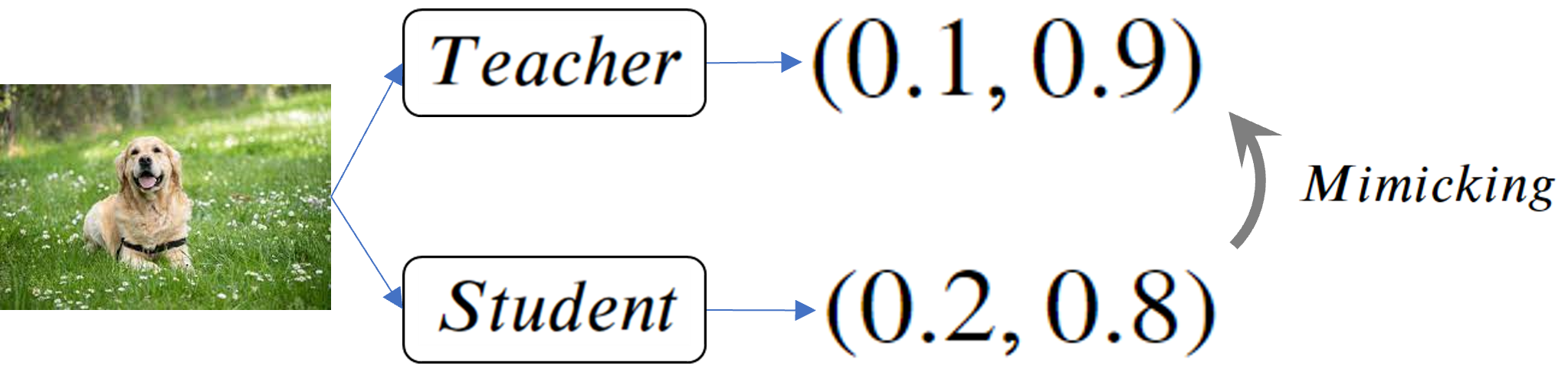}}
    \end{minipage}
    \medskip 
    \begin{minipage}[b]{.31\linewidth}
        \centering
        \subfloat[][cGAN,horse2zebra]{\label{fig:horse_zebra}
        \includegraphics[scale=0.25]{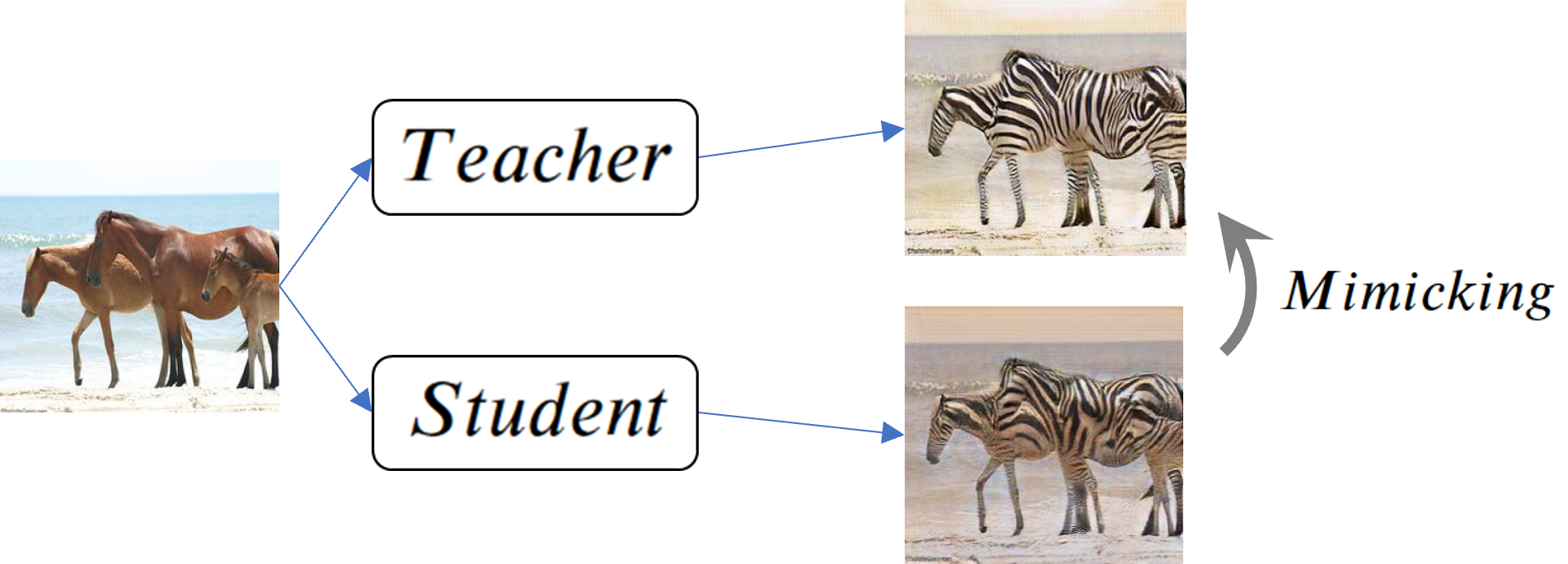}}
    \end{minipage}
    \medskip
    \begin{minipage}[b]{.31\linewidth}
        \centering
        \subfloat[][uncGAN,face generation]{\label{fig:man_woman}
        \includegraphics[scale=0.25]{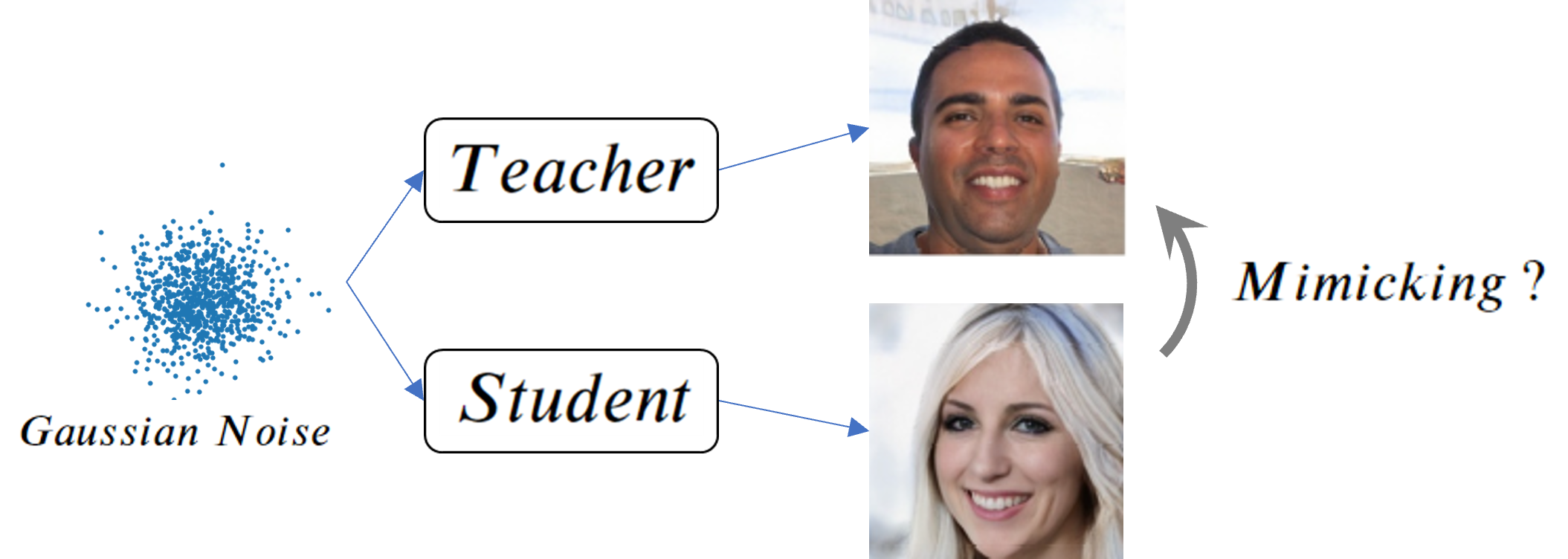}}
    \end{minipage}
    
    \vspace{-10pt}
    \caption{Output discrepancy issue. For the classification task in (a), teacher and student naturally have similar output due to the label supervision. For the conditional GAN such as image-to-image translation in (b), teacher and student also have similar outputs because the input image imposes strong constraints on the output. However, for unconditional generation in (c), teacher and student may produce two images with totally different semantic features. In this condition, distillation is no longer meaningful and cannot bring gains to the student.}
    \label{fig:teaser}
    \vspace{-15pt}
\end{figure}

We find that the main difficulty of uncGAN distillation lies in the \textit{output discrepancy} between the teacher and student model. %, \ie, the student and teacher have totally different outputs given the same input.
An example is shown in Fig.~\ref{fig:teaser}.
In fact, the implicit prerequisite of KD is that teacher and student should have similar outputs for the same input, otherwise the mimicking supervision is no longer meaningful.
This prerequisite is easier to be satisfied in most of cGANs, because the output space of cGANs can be narrowed down by the given conditional input, especially when the condition is strong~\cite{cyclegan,pix2pix}.
Take the horse$\to$zebra task as an example. An input horse image determines which region should be added with zebra stripes and which region is background that should not be changed. 
Two generated images in cGAN may differ in some low-level details such as the shape of zebra stripes, but would largely resemble in their structure.
Unlike cGANs, as shown in our experiments, it is impossible for an uncGAN student with random initialization to learn similar mapping function to the teacher, even though we leverage distillation loss to enforce the agreement between the outputs of two models.

% \if 0
% In this work, we comprehensively investigate the distillation of unconditional GANs.
% %
% Among various uncGANs, style-like GANs, such as StyleGANv1~\cite{stylegan}, StyleGANv2~\cite{stylegan2}, StyleGANv3~\cite{stylegan3}, StyleGAN-ada~\cite{stylegan_ada} and StyleGAN-xl~\cite{stylegan_xl} bring the community impressive results when they are first proposed. 
% %
% Style-like GANs are also the mainstream methods of unconditional GANs.
% %
% Hence, in this paper, we put our focus on the distilling of style-like GANs.
% \fi

To study the aforementioned output discrepancy problem, we focus our attention on the StyleGAN family, \eg,~StyleGAN2~\cite{stylegan2} and StyleGAN3~\cite{stylegan3}, which is one of the most applied unconditional GANs in various downstream tasks~\cite{wang2021HFGI,explaining_in_style,alaluf2021restyle}.
We carefully examine each component of the StyleGAN-like student model through comparative experiments. We identify that the mapping network plays a crucial role in deciding the semantic information of the generated images. Based on this finding, we propose a simple yet effective initialization strategy for the student model, \ie, inheriting the weights from the teacher mapping network and keeping the remaining convolutional layers randomly initialized. Such initialization strategy can work well even in heterogeneous distillation where the student architecture is obtained by neural architecture search (NAS) or manual design, and is totally different from the teacher model.

After resolving the output discrepancy problem, we further design an effective mimicking objective tailored for uncGAN distillation.
As opposed to most of existing GAN distillation approaches that merely transfer the knowledge within a single image, we propose a novel latent-direction-based relational loss to fully exploit the rich relational knowledge between different images.
Specifically, we exploit the good linear separability property of StyleGAN-like models in latent space and augment each latent code $\mathrm{w}$ by moving it along certain direction such that the resulting image only differs in a \emph{single} semantic factor.
Then, we compute the similarity matrix between original images and augmented images and take it as the dark knowledge to be mimicked by the student.
The latent-direction-based augmentation disentangles various semantic factors and makes the learning of each factor easier, thus yielding better distillation performance.

Our \textbf{contributions} are summarized as follows:
\textbf{1)} To the best of our knowledge, this is the first work that uncovers the \textit{output discrepancy} issue in StyleGAN distillation. Through carefully designed comparative experiments, we identify that the mapping network is the determining factor to ensure output consistency.
\textbf{2)} We propose a concise yet effective initialization strategy for the student to resolve the output discrepancy problem, demonstrating significant gains upon conventional uncGAN distillation.
\textbf{3)} We further propose a latent-direction-based distillation loss to learn the rich relational knowledge between different images, and achieve state-of-the-art results in StyleGAN2/3 distillation, outperforming the existing state-of-the-art CAGAN~\cite{CAGAN} by a large margin.

% !TEX root = ../iclr2022_conference.tex

\section{Related Work}
\label{sec:rel_work}

\textbf{GAN Compression.}
We highlight a few recent methods among many GAN compression methods~\cite{co_evo,biggan_distill,portable,ganslimming,gan_compression,slimmable}.
GAN Slimming~\cite{ganslimming} integrates model distillation, channel pruning and quantization into a unified framework.
GAN Compression~\cite{gan_compression} searches a compact student architecture via NAS, and then forces the student to mimic the intermediate outputs and synthesized results of the teacher simultaneously. 
A common characteristic shared by these works is that they all focus on the cGANs such as pix2pixGAN~\cite{pix2pix} and CycleGAN~\cite{cyclegan}.

Aguinaldo's work~\cite{GANKD} focuses on the uncGANs (DCGAN) distillation on low-resolution (32×32) datasets, where the easy setting makes it possible to solve the output discrepancy by adding L1 loss.
Our work explores the distillation of StyleGAN-like models on high resolution (256/1024) images. In this case, output discrepancy issue becomes much more challenging.
The more recent MobileStyleGAN~\cite{mobilestylegan} and Content-Aware GAN compression (CAGAN)~\cite{CAGAN} shift the attention to styleGANs. 
MobileStyleGAN compresses the model by mimicking the wavelet transformation of generated images.
CAGAN estimates the contribution of each channel to the generated faces and eliminates channels with little contribution.
Subsequently, the pruned model inherits the parameters from the original network for both mapping network and convolutional layers, and are finetuned with adversarial loss and distillation loss afterwards.
Though CAGAN involves the compression of uncGAN, it bypasses the issues of model heterogeneity between the teacher and student model by allowing the student to inherit the parameters. Such an requirement assumes the student to inherit the main structure of the teachers too despite pruning.
As will be shown in the experiments, the performance of CAGAN greatly degrades in heterogeneous distillation. 
The proposed mimicking loss cannot guarantee the student to learn a similar mapping as the teacher.
Moreover, we find that the content-aware pruning strategy in CAGAN is not an optimal solution for student initialization.
With our proposed initialization strategy, the student model does not need to inherit any weights from convolutional layers of the teacher but achieves better results.

\noindent
\textbf{Knowledge Distillation.}
 KD~\cite{KD} is originally proposed to achieve model compression~\cite{model_compression} for image classification, whose target is to transfer the dark knowledge from one or multiple cumbersome networks (teacher) to a small compact network (student).
Vanilla KD~\cite{KD} proposes to match the outputs of two classifiers by minimizing the KL-divergence of the softened output logits.
Besides the output logits, other intermediate outputs such as feature maps~\cite{fitnets}, attention maps~\cite{AT,SAD}, Gram matrices~\cite{FSP}, pre-activations~\cite{overhaul}, relation~\cite{cc,simi} and self-supervision signals~\cite{CRD,SSKD} can also serve as the dark knowledge.
However, it should be careful when adapting KD from classification tasks to generation tasks.
The output consistency prerequisite is naturally satisfied in image classification since the supervision of labels guarantees different models to converge to similar mappings.
As discussed in Sec.~\ref{sec:intro}, the consistency prerequisite does not naturally hold for uncGANs.
Therefore, a special distillation technique tailored for uncGANs is required to cope with the output discrepancy problem.

\noindent\textbf{StyleGAN Linear Property.}
As shown in StyleGAN~\cite{stylegan}, for a well-trained model, the $w$ latent space consists of linear subspaces. 
It should be possible to find direction vectors that consistently correspond to individual factors of variation.
Recently, some works~\cite{gan_space,hessian_penalty,sefa,unsup_discovery} have been conducted to find these meaningful directions.
Among them, SeFa~\cite{sefa} finds the latent directions by computing the eigenvalues of the transformation matrix in the ModConv~\cite{AdaIN} layer.
We adopt it in our latent-direction-based loss due to its fast computation and high performance.
A recent work StyleAlign~\cite{stylealign} provides a thorough analysis about the property of StyleGAN latent space.
It finds that the latent directions control similar semantic factors for two aligned models even they work on very different domains.
This finding aligns with our observation that the mapping network plays a vital roles in determining the semantics of generated images.

% !TEX root = ../iclr2022_conference.tex

\section{Methodology}

\subsection{Preliminaries}
\label{sec:preliminary}

\textbf{StyleGAN.}
There are two modules in StyleGAN-like models~\cite{stylegan,stylegan2,stylegan3}, \ie, a mapping network $S(\cdot)$ that maps Gaussian noise $z$ to the style vector $w$ and a convolution backbone $C(\cdot)$ that takes $w$ as input and generates images.
The style vector $w$ is fed into the backbone $C(\cdot)$ through the modulated convolution (ModConv) layer~\cite{AdaIN,stylegan2}.
StyleGAN allows the use of different $w$ vectors in different ModConv layers.
The image generation process can be formulated as:
\begin{equation}
    G(z_1,z_2,\cdots,z_L) = C(w_1,w_2,\cdots,w_L) = C(S(z_1), S(z_2), \cdots, S(z_L)),
\end{equation}
where $L$ is the number of ModConv layers in the backbone, and the $i$-th ModConv layer uses $w_i$ that comes from $z_i$.
We define the output consistency condition as:
\begin{equation}
\label{eq:consistency1}
    G_s(z_1,z_2,\cdots,z_L) = G_t(z_1,z_2,\cdots,z_L),
\end{equation}
where the $s$ and $t$ represent student and teacher, respectively. %\yuenan{the output consistency condition is too strict, even using our method, the output of student and teacher cannot be exactly the same} 
Equation~\ref{eq:consistency1} suggests that the generated images of two models should be the same if they use the same $z$ at corresponding layers.

\noindent
\textbf{StyleGAN Compression.}
A typical StyleGAN compression approach~\cite{CAGAN} contains two steps, \ie, pruning and finetuning.
In the pruning stage, unimportant / unnecessary channels will be removed according to some heuristics~\cite{low_act,li2017pruning,he2018soft,CAGAN}. %, such as channels with low activation~\cite{low_act} or channels that make little contribution to the output~\cite{CAGAN}.
Note that pruning is only applied to the convolution backbone $C(\cdot)$ and the mapping network $S(\cdot)$ is kept \textit{unchanged}.
The pruned model will inherit the well-trained weights from the original model for both the mapping network and the convolution backbone~\cite{CAGAN}.
In the finetuning stage, besides the normal adversarial loss, the pruned model is also required to mimic the original model's output to compensate the performance degradation brought by channel reduction.
A typical mimicking loss includes RGB loss and LPIPS loss~\cite{lpips}:
\begin{equation}
    \mathcal{L}_{\mathrm{rgb}} = ||G_s(z) - G_t(z)||_1, \mathcal{L}_{\mathrm{lpips}} = ||F(G_s(z))-F(G_t(z))||_1,
\end{equation}
where $F$ is a well-trained frozen network that computes the perceptual distance between two images.
$L_{\mathrm{rgb}}$ and $L_{\mathrm{lpips}}$ require that the generated image of student should be close to that of teacher in RGB space and perceptual space, respectively.
The final loss function in the finetuning stage is:
\begin{equation}
\label{eq:finetune}
    \mathcal{L} = \lambda_{\mathrm{GAN}}\mathcal{L}_{\mathrm{GAN}} + \lambda_{\mathrm{rgb}}\mathcal{L}_{\mathrm{rgb}} + \lambda_{\mathrm{lpips}}\mathcal{L}_{\mathrm{lpips}},
\end{equation}
where $\lambda_*$ is the loss weight of each item.

\subsection{Framework Overview of Unconditional GAN Distillation}
\label{sec:prerequisite}

Knowledge distillation is a common strategy that can bring improvements in classification tasks.
However, in generation tasks, its prerequisite, namely the student and teacher having consistent outputs for the same input, is rarely mentioned.
In the absence of this prerequisite, the influence of mimicking losses on the training of student remains largely unknown.
Here, we hypothesize that RGB or LPIPS loss is not compatible with GAN loss when the output discrepancy occurs and distillation will also bring no benefit to the student.
We examine this hypothesis both qualitatively and quantitatively.

\begin{figure}[t]
    \begin{minipage}[b]{.58\linewidth}
        \centering
        \subfloat[][Qualitative effects of RGB/LPIPS losses.]{\label{fig:LPIPS_RGB_change}
        \includegraphics[scale=0.26]{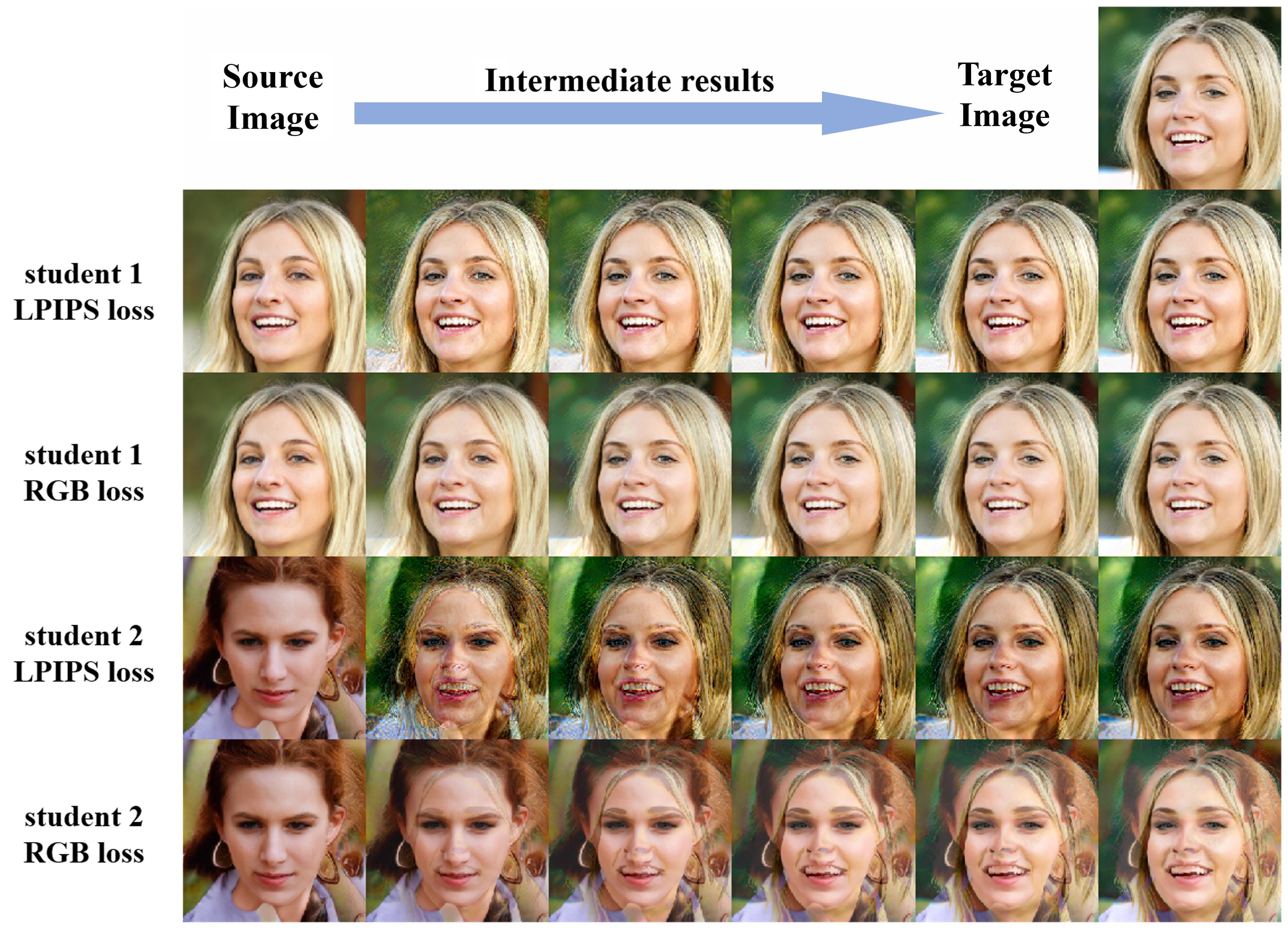}}
    \end{minipage}
    \medskip 
    \hspace{-15pt}
    \begin{minipage}[b]{.5\linewidth}
        \centering
        \subfloat[][Grad cosine between GAN and RGB loss.]{\label{fig:GAN_RGB_cosine}
        \includegraphics[scale=0.16]{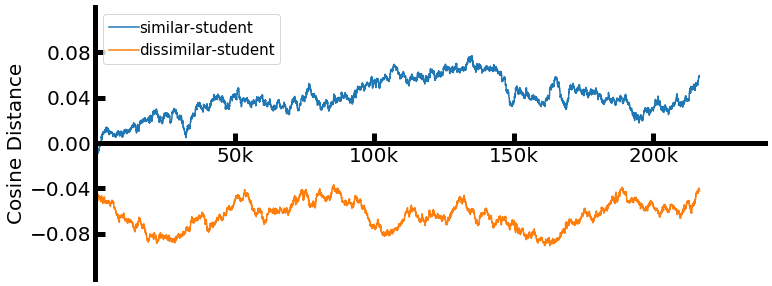}}\\
        \subfloat[][Grad cosine between GAN and LPIPS loss]{\label{fig:GAN_LPIPS_cosine}
        \includegraphics[scale=0.16]{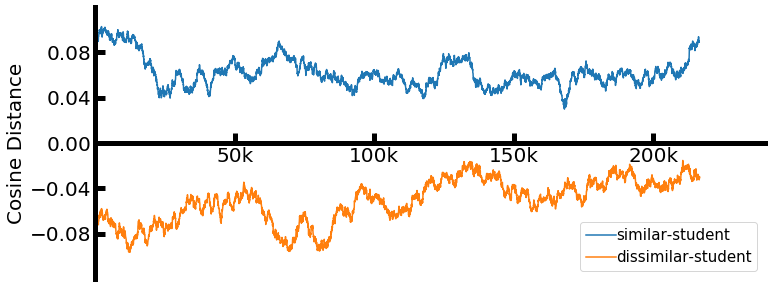}}
    \end{minipage} %\par
    \vspace{-7pt}
    \caption{(a) Student-1 has similar outputs with teacher and Student-2 has different outputs. The image in the top right corner is the teacher output. We demonstrate the intermediate results to show how RGB/LPIPS loss influences the image generation of student. (b)(c) Cosine distance between the gradient of GAN loss and RGB/LPIPS loss. The $x-$axis denotes training steps. For similar student, RGB/LPIPS loss is cooperating with GAN loss. For dissimilar student, RGB/LPIPS loss is competing with GAN loss.}
    \label{fig:prerequisite}
    \vspace{-20pt}
\end{figure}

Note that the three losses in Eq.~\ref{eq:finetune} serve different roles. 
$\mathcal{L}_{\mathrm{GAN}}$ requires the student to generate realistic images while $\mathcal{L}_{\mathrm{rgb}}$ and $\mathcal{L}_{\mathrm{lpips}}$ encourage similarity between the generated images by student and those of the teacher.
Intuitively, if the generated image of the student is totally different from that of teacher for the same input, $\mathcal{L}_{\mathrm{rgb}}$ and $\mathcal{L}_{\mathrm{lpips}}$ will result in images that are slightly closer to teacher but with much less realism.
To examine this hypothesis, we remove the GAN loss in Eq.~\ref{eq:finetune} and keep only RGB or LPIPS loss. 
We also cut off the gradient backward path between the student generator and generated images. In this condition, the gradient of RGB/LPIPS loss directly works on the images.
The change of synthesized images reflects how RGB/LPIPS loss influences the generation process.
We select two student models, \ie, student-1 that has similar output with teacher for the same input and student-2 that has totally different outputs from teacher.
Two students have identical architectures. The mapping network of student-1 inherits from teacher and the mapping network of student-2 is randomly initialized.
The effects of RGB/LPIPS loss are shown in Fig.~\ref{fig:LPIPS_RGB_change}. 
We can find that the intermediate results are a mixup of source and target images to some extent.
If the source image is in the neighbourhood of the target image (1st and 2nd rows), the intermediate results are still perceptually realistic.
However, if the source image is totally different from the target image (3rd and 4th rows), the intermediate results are no longer realistic.
Though RGB and LPIPS losses are reducing the distance between source and target images, they cannot guarantee a smooth and face-like interpolation in the dissimilar setting.
And this unrealistic intermediate results naturally contradict with GAN loss.

From quantitative perspective, we wish to prove that RGB/LPIPS loss is not compatible with GAN loss in the heterogeneous setting by gradient analysis.
In the training process, for each batch, we perform backward propagation for GAN loss, RGB loss and LPIPS loss, respectively, and obtain three gradients of these losses.
We then compute the cosine distance between GAN gradient and RGB/LPIPS gradients.
As shown in Fig.~\ref{fig:GAN_RGB_cosine} and Fig.~\ref{fig:GAN_LPIPS_cosine}, the cosine distance between GAN gradient and RGB/LPIPS gradients of dissimilar student is always negative, suggesting that RGB/LPIPS gradients are competing with GAN gradients.
On the contrary, the cosine distance of similar student is positive, indicating that the distillation loss is driving the model in the same direction as the adversarial loss.
Our analysis above suggests that distillation is not beneficial in heterogeneous setting. Having similar outputs for the same input $z$ is the prerequisite for uncGAN distillation.

\subsection{Effect of the Mapping Network}
\label{sec:style_module}

As we will show in the experiments, if the student is randomly initialized, it cannot learn consistent outputs as teacher even though we leverage RGB/LPIPS loss to force the agreement between the outputs of two models.
We hypothesize that the mapping network $S(z)$ plays a key role in determining whether two models can have consistent outputs.
If the gap between mapping networks of student and teacher is too large, it is hard for the student to learn outputs consistent with the teacher.
This hypothesis comes from the following motivation.

Suppose the student has a different mapping network from the teacher and the consistency condition (Eq.~\ref{eq:consistency1}) is still satisfied. 
Our goal is to derive a contradiction.
For the convenience of the following discussion, we define:
\begin{equation}
    G(z_1, z_2;k)=C(w_1,w_2;k) = C(w_1,\cdots,w_1,w_2,w_1,\cdots,w_1),
\end{equation}
where all the ModConv layers use $w_1$ except that the $k$-th layer uses $w_2$.
The consistency condition of Eq.~\ref{eq:consistency1} requires that:
\begin{equation}
\label{eq:consistency2}
    G_s(z_1,z_2;k) = G_t(z_1,z_2;k), 1\leq k\leq L.
\end{equation}
As shown in StyleGAN~\cite{stylegan}, for a well-trained model, it should be possible to find direction vectors that consistently correspond to individual factors of variation.
An example is shown in appendix~\ref{fig:sefa}. 
Some individual semantic factors such as pose, glasses and hair color can be controlled by moving the style vector $w$ of certain layer along a certain direction.
Suppose the direction $p$ at $k$-th layer controls the hair color of the generated face.
The only difference between $C_t(w_0,w_1+p;k)$ and $C_t(w_0,w_1;k)$ is that they are the same faces with different hair colors.
The movement of $w$ from $w_1$ to $w_1+p$ corresponds to a consecutive change of hair color of the generated face.
If we map the $w$ back to the noise space:
\begin{equation}
    z_1 = S_t^{-1}(w_1), \;\;\; z_2 = S_t^{-1}(w_1+p),
\end{equation}
obviously, the line segment in $w$ space corresponds to a curve in $z$ space with two end points $z_1$ and $z_2$ due to the nonlinearity of $S_t(z)$.
We denote this curve as $\overset{\frown}{z_1z_2}$.
Then $\{G_t(z_0,z;k)|z\in\overset{\frown}{z_1z_2}\}$ represents a cluster of faces with different hair colors.
According to the consistency constraint, $\{G_s(z_0,z;k)|z\in\overset{\frown}{z_1z_2}\}$ should be the same cluster as $\{G_t(z_0,z;k)|z\in\overset{\frown}{z_1z_2}\}$.
We feed $z_0, z\in\overset{\frown}{z_1z_2}$ into the student mapping network $S_s(\cdot)$:
\begin{equation}
    w_0^{\prime}=S_s(z_0), \;\;\; \overset{\frown}{w_1^{\prime}w_2^{\prime}}=S_s(\overset{\frown}{z_1z_2}).
\end{equation}
Since $S_s(\cdot)$ is different from and independent of $S_t(\cdot)$, the result $\overset{\frown}{w_1^{\prime}w_2^{\prime}}$ is still a curve.
Thus, the semantic factor of hair color in student model is controlled by a complex curve in $w$ space, which contradicts the property of StyleGAN that various semantic factors are decoupled well in $w$ space. 
Hence, having different mapping networks and consistency condition cannot hold simultaneously.

% %
% We name this kind of direction as latent direction.
% %
% Suppose that student has different mapping network from teacher and two models generate consistent output for the same input noise $z$.
% %
% If the latent direction of teacher is mapped back to noise space and then be remapped to student latent space, considering the nonlinearity and large gap between two mapping networks, the resulting segment in student latent space is no longer linear.
% %
% In other words, the semantic factor that can be controlled by linear movement in teacher latent space now is controlled by non-linear movement in student latent space, which contradicts the good linear separability of StyleGAN2.
% %
% Hence, we conjecture that a large gap between teacher and student mapping networks is not compatible with the consistency condition in Eq.~\ref{eq:consistency1}.

\begin{figure*}[t]
	\centering
	\includegraphics[scale=0.35]{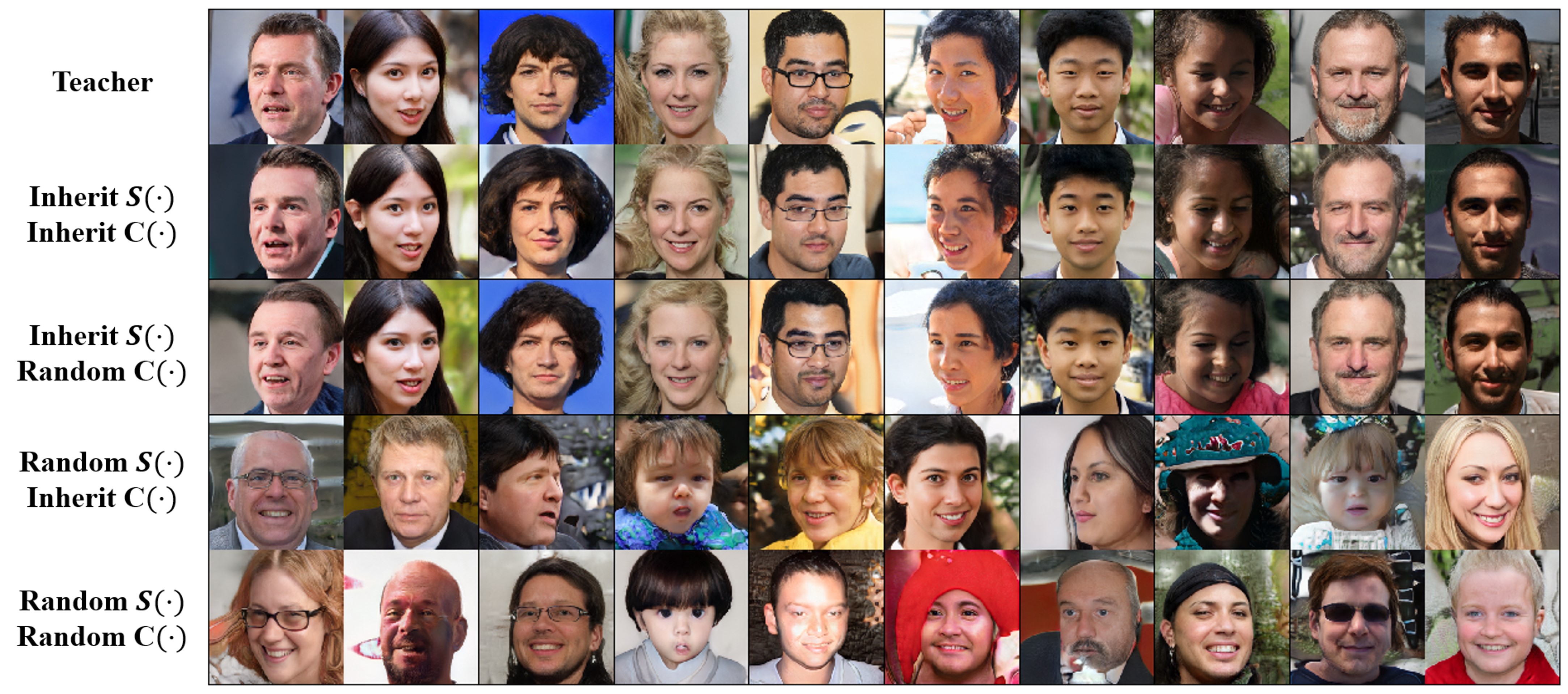}
	\vspace{-5pt}
	\caption{The mapping network $S(\cdot)$ determines whether the student can learn from the teacher's output.}
	\label{fig:style_module}
	\vspace{-10pt}
\end{figure*}

We further conduct experiments to examine our hypothesis.
Specifically, we select four students according to whether the mapping network is from teacher or not and whether the convolution is from teacher or not.
We use GAN loss, RGB loss and LPIPS loss to train these models.
The mapping network and convolution are updated together.
The results are shown in Fig.~\ref{fig:style_module}.
The student that inherits weights from the teacher's mapping network can learn a mapping that aligns well with the teacher's output, no matter how the convolution $C(\cdot)$ is initialized.
However, for the student whose mapping network is randomly initialized, there are no meaningful connections between student's and teacher's outputs.
The analysis above clearly shows that the output consistency between student and teacher is determined by the mapping network.

\subsection{Mapping Network Consistency in GAN Distillation}
\label{sec:two_stage}

We have shown that the consistency between student and teacher outputs is the prerequisite of the distillation, and the mapping network determines whether two generators can have consistent outputs.
Hence, to make distillation meaningful, it is necessary to impose extra constraints to guarantee the consistency between two mapping networks.

The simplest way is to keep the architecture of the mapping network unchanged and inherit teacher's parameters directly.
In fact, the parameters and FLOPs of the mapping network account for only 7.5\% and 0.005\% of the convolution backbone.
Preserving the mapping network is thus feasible in practice.

If there is a strong demand on the compression of the mapping network, one can perform a two-stage training to ensure a small gap between student and teacher mapping networks.
In the first stage, the student mapping network is forced to mimic outputs of the teacher mapping network:
\begin{equation}
\label{eq:two_stage}
    \mathcal{L} = \mathbb{E}_{z\sim\mathcal{N}(0,1)} D(S_s(z), S_t(z)),
\end{equation}
where $D(\cdot,\cdot)$ is a distance metric.
Considering $S_t(\cdot)$ and $S_s(\cdot)$ are both shallow MLPs, the training cost of this stage is negligible (0.59\% of the normal GAN training in the second stage).
In the second stage, the mapping network and generator backbone are finetuned together using the loss in Eq.~\ref{eq:finetune}.
We will explore the effects of compressing the mapping network in Sec.~\ref{sec:ablation}.

\subsection{Latent-Direction-Based Relation Distillation}
\label{sec:relation_loss}

Under the premise that consistency condition is satisfied, we further propose to incorporate relation mimicking into GAN distillation.
Conventional relation-based distillation~\cite{simi} in classification tasks computes feature similarity matrices using the samples in a minibatch.
Here, we tailor it to better cater to StyleGAN.

Specifically, for a given teacher model, we compute its meaningful latent directions (LD) that control a single semantic factor and store them in a dictionary $\{d_1, d_2,\cdots,d_m\}$.
Note that the latent direction is related to a specific layer. 
For example, if $d_i$ is computed in $k$-layer, then only $C_t(w,w+d_i;k)$ has single semantic factor difference with $C_t(w)$. $C_t(w,w+d_i;j)_{j\neq k}$ does not has this property.
In the training stage, we feed a batch of noise $\{z_i\}_{i=1:N}$ into the mapping network and obtain $\{w_i\}_{i=1:N}$.
For each $w_i$ we randomly sample a latent direction $d$ from the dictionary.
Thus, $C_t(w_i)$ and $C_t(w_i,w_i+\alpha d;k)$ ($k$ is the layer related to $d$) are two images with single semantic factor difference with $\alpha$ controls the moving distance.
We denote the intermediate features of $C_t(w_i)$ and $C_t(w_i,w_i+\alpha d;k)$ as $f_i$ and $f_i'$, respectively.
Then the similarity matrix $M$ between original view and augmentation view can be computed as $A_{i,j} = f_i\cdot f_j'$. %, where $\cdot$ denotes the inner product.
%
%We can directly minimize the distance between the similarity matrix of teacher and student using L2 loss.
%
We then convert the similarity into probability via the softmax operation and minimize the distance using KL-divergence loss:
\begin{equation}
    M_{i,j} = \frac{\exp(A_{i,j})}{\sum_{k=1}^N \exp(A_{i,k})},  \;\;\;  \mathcal{L}_{\mathrm{LD}} = -\sum_{i,j} M_{i,j}^t \log M_{i,j}^s.
\end{equation}
The final learning objective is the combination of Eq.~\ref{eq:finetune} and $\mathcal{L}_{\mathrm{LD}}$. %We will compare these two loss in the ablation study.

% !TEX root = ../iclr2022_conference.tex

\begin{table}[t]
    % \vspace{-5pt}
    \centering
    \caption{Effect of initialization. Surprisingly, we find that inheriting only mapping network is the best solution.}
    \vspace{5pt}
    \begin{tabular}{cccc}
    \toprule
    \multirow{2}*{\shortstack{mapping network\\ Initialization}} & 
    \multirow{2}*{\shortstack{Convolution\\Initialization}} & 
    \multirow{2}*{\shortstack{Mimicking Loss}} & 
    \multirow{2}*{Student FID} \\
     ~ & ~ & ~ & ~  \\
    \midrule
     \multirow{3}*{random} & \multirow{3}*{random} & No Mimic & 10.92 \\
     ~ & ~ & RGB & 10.78 \\
    ~ & ~ & RGB + LPIPS & 11.27 \\
    \midrule
    \multirow{2}*{random} & \multirow{2}*{inherit} & RGB & 10.81 \\
    ~ & ~ & RGB+LPIPS & 10.88\\
    \midrule
     \multirow{4}*{inherit} & \multirow{4}*{inherit} & No Mimic & 10.54 \\
     ~ & ~ & RGB & 9.41 \\
     ~ & ~ & RGB + LPIPS & 8.61 \\
     ~ & ~ & RGB + LPIPS + LD & 8.45 \\
    \midrule
    \multirow{3}*{inherit} & \multirow{3}*{random} & RGB & 9.42 \\
     ~ & ~ & RGB + LPIPS & 8.23 \\
     ~ & ~ & RGB + LPIPS + LD & \textbf{7.94} \\
    \bottomrule
    \end{tabular}
    \label{tab:ablation_init}
    \vspace{-5pt}
\end{table}

%Preserving mapping network architecture and random initialization leads to the worse result. Compressing mapping network and using two-stage strategy can improvement performance. Preserving architecture and inheriting teacher weights is the best solution. Considering mapping network FLOPs accounts only 0.005\% for convolution FLOPs, the loss of FID outweighs the gain of global FLOPs saving.

\section{Experiments}
\label{sec:label}

We conduct experiments mainly on StyleGAN2/3 since they are the most powerful unconditional GANs so far. We use the FFHQ~\cite{stylegan} and LSUN church~\cite{lsun} datasets.
We adopt Fr\'echet Inception Distance (FID), Perceptual Path Length (PPL)~\cite{stylegan} and  PSNR~\cite{CAGAN} between real and projected images as evaluation metrics.
More qualitative results are shown in the appendix~\ref{sec:qualitative}.

For the ablation study in Sec.~\ref{sec:ablation}, we train the models on resolution 256$\times$256 and use a smaller batch size of 8 to save the computation cost.
For the comparison with state-of-the-art methods in Sec.~\ref{sec:benchmark}, we train the models on both resolutions of 256$\times$256 and 1024$\times$1024.
We also use a batch size of 16 that is the same as CAGAN~\cite{CAGAN} to ensure a fair comparison.

\subsection{Ablation Study}
\label{sec:ablation}

\textbf{The Initialization of the Student Model.}
Previous works usually treat StyleGAN2 as an integral module and initialize the mapping network and convolution backbone in the same way (from scratch or inherits teacher parameters).
Based on our analysis in Sec.~\ref{sec:style_module} that the mapping network plays a key role in determining the semantics of generated images, here we separate the mapping network $S(z)$ from the convolution backbone $C(w)$ and test three initialization strategies: 1) both $S(z)$ and $C(w)$ are randomly initialized, 2) both $S(z)$ and $C(w)$ are initialized with teacher weights, 3) only $S(z)$ inherits teacher weights and $C(w)$ is randomly initialized.

The results are shown in Table~\ref{tab:ablation_init}.
For the setting where $S(z)$ and $C(w)$ are both randomly initialized, RGB loss can only bring marginal improvement. 
RGB+LPIPS even performs worse than No-Mimic, indicating that distillation cannot work well when output discrepancy occurs.
If $S(z)$ and $C(w)$ both inherit teacher weights, the mimicking loss can achieve 1-2 FID improvement.
To explore the effect of the mapping network, we also try inheriting only $S(z)$ and surprisingly find that this initialization obtains the best result.
And loading $C(w)$ hampers the performance of distillation.
This result contradicts with the conclusion in CAGAN.
It shows that the general pruning strategy, \ie, determining which channels should be removed, is not important.
Randomly initialization of convolution layers is the optimal solution if the mapping network is kept.

\noindent
\textbf{The Effects of Mapping Network Compression.}
We conduct experiments to investigate how to deal with the mapping network in StyleGAN-like models compression.
Specifically, we consider three settings: 1) student has the same mapping network architecture as teacher but with random initialization, 2) student mapping network has a different architecture and uses the two-stage training strategy, 3) student has the same architecture and inherits weights from the teacher.
For all the settings, the convolution backbones are randomly initialized.
For the two-stage setting, we also explore how the architecture of the mapping network and mimicking loss in Eq.~\ref{eq:two_stage} affect the final performance.
To emphasize the importance of the mapping network, we also list the average L1 distance between $S_s(z)$ and $S_t(z)$ before entering the normal GAN training stage. 

\begin{table}[t]
    \centering
    \caption{How to deal with the mapping network in StyleGAN2 distillation. The mapping network is comprised of MLPs. The numbers inside and outside the ``[]'' is the number of channels in each layer and the number of layers, respectively. The mapping network of the teacher is [512]*8. The FLOPs saving is computed with regard to the total FLOPs (the mapping network and convolution layers).}
    % \vspace{-5pt}
    \begin{tabular}{cccccc}
    \toprule
    \multirow{2}*{\shortstack{Setting}} & 
    \multirow{2}*{\shortstack{mapping network\\Architecture}} & 
    \multirow{2}*{\shortstack{FLOPs\\Saving}} & 
    \multirow{2}*{$D(\cdot, \cdot)$} & 
    \multirow{2}*{$||S_s(z)-S_t(z)||_1$} & 
    \multirow{2}*{Student FID} \\
    ~ & ~ & ~ & ~ & ~ & ~ \\
    \midrule
    Random Initialization & [512]*8 & 0\% & N/A & 1.027 & 11.78 \\
    \midrule
    \multirow{5}{*}{Two-Stage} & [512]*8 & 0\% & L1 & 0.156 & 9.69 \\
    ~ & [512]*8 & 0\% & L2 & 0.260 & 10.80 \\
    ~ & [512]*5 & 0.0019\% & L1 & 0.197 & 10.38 \\
    ~ & [390]*7+[512] & 0.0019\% & L1 & 0.210 & 10.55 \\
    ~ & [256]*7+[512] & 0.0034\% & L1 & 0.245 & 10.86 \\
    \midrule
    Inheriting & [512]*8 & 0\% & N/A & 0 & \textbf{8.30} \\
    \bottomrule
    \end{tabular}
    % \vspace{-10pt}
    \label{tab:ablation_style_module}
    \vspace{-10pt}
\end{table}

The results are shown in Table.~\ref{tab:ablation_style_module}.
`Random Initialization' obtains the worst FID because the output discrepancy makes the distillation ineffective.
The `Two-Stage' strategy improves the results by narrowing the gap between $S_s(z)$ and $S_t(z)$.
From several two-stage settings, we can find that L1 is a better mimicking loss than L2 and reducing the number of layers is better than reducing the number of channels in each layer.
It is also worth noting that there is a strong positive correlation between $|S_s(z)-S_t(z)|$ and FID, indicating that the gap between $S_s(z)$ and $S_t(z)$ determines the output consistency and further determines the influence of distillation.
Though the two-stage strategy brings performance gains, there is still a large gap between it and the `Inheriting' variant.
Thus, we conclude that the modification to the mapping network will greatly harm the final performance and the two-stage strategy can only mitigate the degradation to a certain degree.
Considering that the scale of the original $S_t(z)$ is negligible compared to the convolution backbone, the best practice in StyleGAN2 compression is to preserve the mapping network architecture and inherit the weights from the teacher mapping network.

\begin{table}[t]
    \renewcommand\arraystretch{1.08}
    \centering
    % \small
    \caption{Comparison with SOTA methods. ``$\downarrow$'' (``$\uparrow$'') denotes the lower (higher) the better. ``$\dagger$'' denotes that the numbers come from CAGAN~\cite{CAGAN}. \textbf{Bold} font denotes the results that outperform CAGAN. ``heter'' denotes the heterogeneous setting where the student is not a subnet of the teacher. Since StyleGAN3 removes the PPL loss in the training stage, we also do not measure the PPL for StyleGAN3. The PSNR (proposed by CAGAN) is a special-designed metric to measure the face projection ability. Thus, we do not measure it for the LSUN church dataset. We compute PSNR using our own implementation and leave the result of GAN slim blank due to the lack of the corresponding checkpoint.}
    \vspace{5pt}
    \begin{tabular}{c|c|c|c|c|c|c|c|c}
    \hline
    Model & Dataset & Reso. & Methods & RAM & FLOPs & FID ($\downarrow$) & PPL ($\downarrow$) & PSNR ($\uparrow$)\\
    \hline
    \multirow{14}{*}{StyleGAN2} & \multirow{11}{*}{FFHQ} & \multirow{7}{*}{256} & Teacher & 30.0M & 45.1B & 4.5 & 0.162 & 34.26\\
    \cline{4-9}
    ~ & ~ & ~ & Baseline & 5.6M & 4.1B & 9.79 & 0.156 & 33.17\\
    ~ & ~ & ~ & GAN slim & - & 5.0B & 12.4$\dagger$ & 0.313$\dagger$ & -\\
    ~ & ~ & ~ & CAGAN & 5.6M & 4.1B & 7.9$\dagger$ & 0.143$\dagger$ & 33.34\\
    ~ & ~ & ~ & \textbf{Ours} & 5.6M & 4.1B & \textbf{7.25} & \textbf{0.135} & \textbf{33.49}\\
    \cline{4-9}
    ~ & ~ & ~ & CAGAN-heter & 3.4M & 2.7B & 13.75 & 0.158 & 33.19\\
    ~ & ~ & ~ & \textbf{Ours-heter} & 3.4M & 2.7B & \textbf{9.96} & \textbf{0.141} & \textbf{33.54}\\
    \cline{3-9}
    ~ & ~ & \multirow{4}{*}{1024} & Teacher & 49.1M & 74.3B & 2.7 & 0.162 & 33.52\\
    \cline{4-9}
    ~ & ~ & ~ & GAN slim & - & 23.9B & 10.1$\dagger$ & 0.211$\dagger$ & -\\
    ~ & ~ & ~ & CAGAN & 9.2M & 7.0B & 7.6$\dagger$ & 0.157$\dagger$ & 32.63\\
    ~ & ~ & ~ & \textbf{Ours} & 9.2M & 7.0B & \textbf{7.19} & \textbf{0.128}& \textbf{32.70}\\
    \cline{2-9}
    ~ & \multirow{3}{*}{\shortstack{LSUN\\Church}} & \multirow{3}{*}{256} & Teacher & 30.0M & 45.1B & 4.92 & 0.168 & N/A\\
    \cline{4-9}
    ~ & ~ & ~ & CAGAN & 5.6M & 4.1B & 8.57 & 0.146 & N/A\\
    ~ & ~ & ~ & \textbf{Ours} & 5.6M & 4.1B & \textbf{7.96} & \textbf{0.136} & N/A\\
    \cline{1-9}
    \multirow{3}{*}{StyleGAN3} & \multirow{3}{*}{FFHQ} & \multirow{3}{*}{256} & Teacher & 30.0M & 45.1B & 4.41 & N/A & 34.30\\
    \cline{4-9}
    ~ & ~ & ~ & CAGAN & 5.6M & 4.1B & 7.75 & N/A & 33.39\\
    ~ & ~ & ~ & \textbf{Ours} & 5.6M & 4.1B & \textbf{7.14} & N/A & \textbf{33.58}\\
    \hline
    \end{tabular}
    \label{tab:comparison}
    \vspace{-10pt}
\end{table}

\subsection{Comparison with State-of-the-Art Methods}
\label{sec:benchmark}

\textbf{Quantitative Results}.
We compare our method with the GAN Slimming ~\cite{ganslimming} and CAGAN~\cite{CAGAN} methods. 
Since our method does not focus on the pruning, we directly adopt the student architecture used in CAGAN, \ie, a network that is the same as teacher but with fewer channels.
We also compare with CAGAN in the heterogeneous setting where the student is not a subnet of the teacher.
Specifically, we modify the kernel size of the second convolution layer in each residual block from 3 to 1, thus inheriting teacher convolution parameters is infeasible.
Since CAGAN did not notice the output discrepancy issue and always initialize the mapping network and convolution backbone in the same way, we assume it does not inherit weights from teacher in the heterogeneous setting.

\begin{figure*}[t]
	\centering
	\includegraphics[scale=0.35]{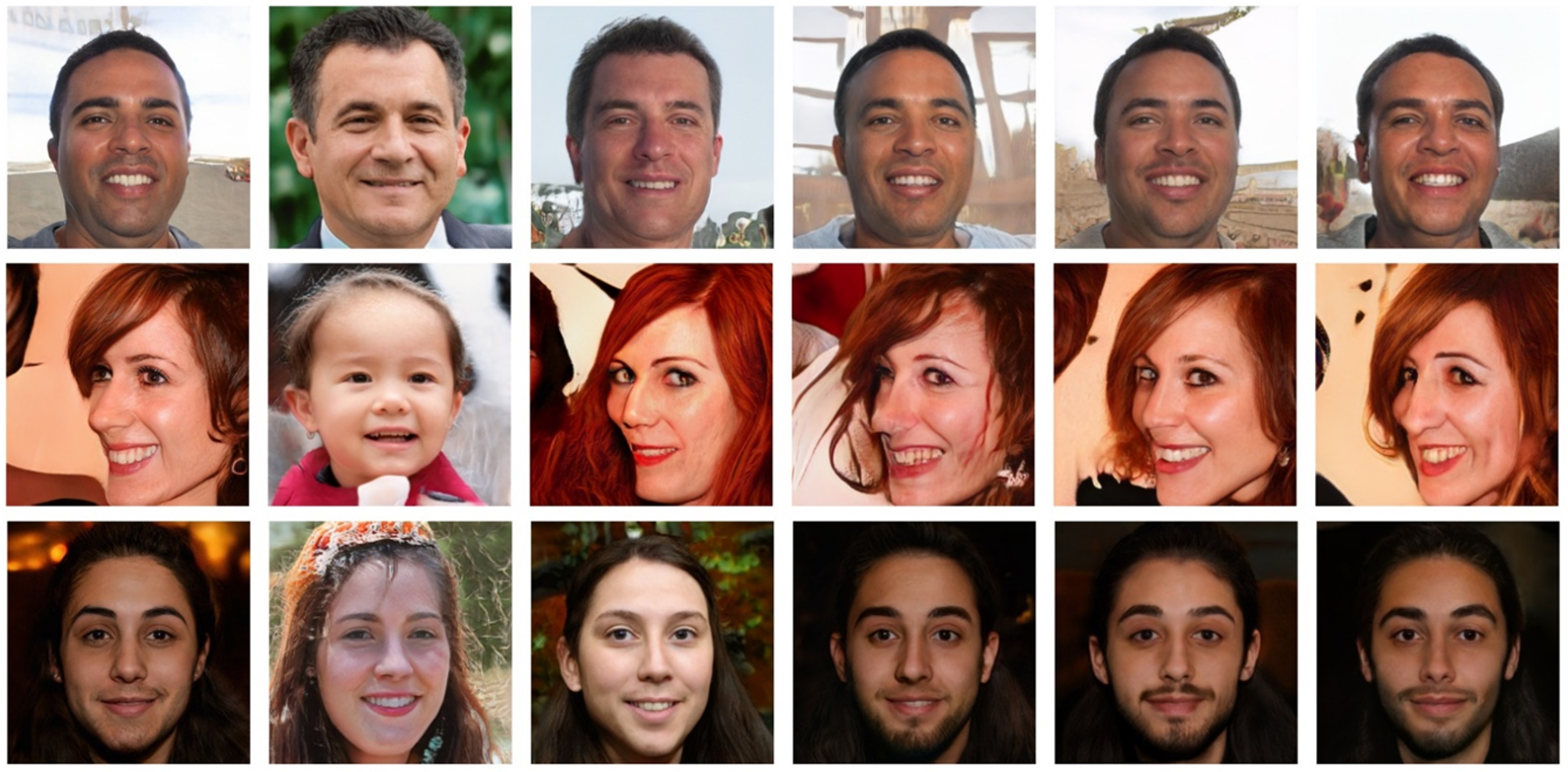}
	\vspace{-5pt}
	\caption{StyleGAN2 synthesized results on FFHQ 256$\times$256.}
	\label{fig:compare_256}
	\vspace{-10pt}
\end{figure*}

The results are shown in Table~\ref{tab:comparison}.
For the distillation of StyleGAN2 on FFHQ dataset, our method outperforms CAGAN on FID by 0.65 and 0.41 on resolution 256$\times$256 and 1024$\times$1024, respectively, showing that our method can generate more realistic images.
Note that these improvements are not marginal considering the images generated by CAGAN are already of high quality.
For the PPL metric that measures the smoothness of latent space, we outperform CAGAN by 6\% (relative improvement) on resolution 256$\times$256. 
The gap is even larger (18.5\%) on resolution 1024$\times$1024.
For PSNR that is related to the image projection ability, our method also surpasses CAGAN, demonstrating that our method can model the face distribution in real world better.
Our superiority is much more significant in the heterogeneous setting, showing that our method can be applied in a more general situation where the student is not necessary to be a subnet of the teacher.
On LSUN Church dataset, our method still achieves better results than CAGAN on both FID and PPL, showing that our method not only handles those well-aligned settings, but also works well in complex outdoor scenes. On StyleGAN3, our method also brings more gains, indicating that the proposed method has good generalization ability in various StyleGAN-like models.

\vspace{5pt}
\noindent
\textbf{Qualitative Results}.
We show StyleGAN2 generation results of FFHQ on resolution 256$\times$256 in Fig.~\ref{fig:compare_256}.
For Two-Stage, we compress the original 8-layer mapping network into 5 layers. The images of each row are generated using the same input noise $z$. Note that all the students are trained with mimicking loss. Random $S_s(z)$ cannot make the student model generate images consistent with the teacher due to the different mapping networks. The Two-Stage method mitigates output discrepancy issue by directly mimicking the mapping network, but there still exist semantic differences from the teacher. Compared to CAGAN, our generated images have fewer artifacts and are more similar to the teacher in various semantic features such as the face color, haircut and expression.

\begin{figure*}[t]
	\centering
	\includegraphics[scale=0.19]{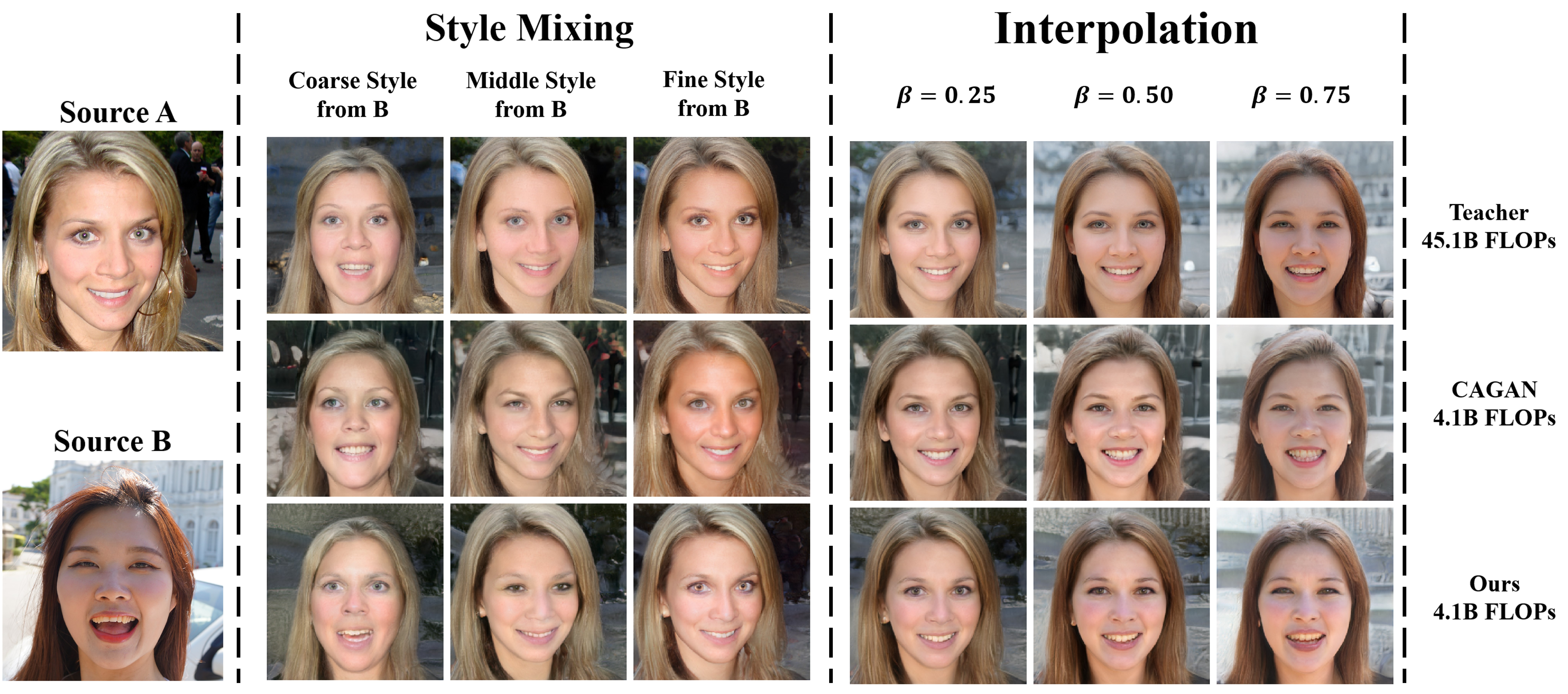}
	\vspace{-5pt}
	\caption{In coarse style mixing, our result corresponds better with source B on the mouth and face shape. In fine style mixing, our result corresponds better with source B on skin color. CAGAN also generates artifacts on hair in middle and fine style cases.}
	\label{fig:editing}
	\vspace{-10pt}
\end{figure*}

\noindent\textbf{Image Editing.}
We demonstrate an image editing case in Fig.~\ref{fig:editing}. Specifically, we apply style mixing and interpolation to the image. The implementation details and more results are shown in the appendix~\ref{fig:editing_supp}.

\section{Conclusion}

In this paper, we uncover the output discrepancy issue in uncGAN distillation.
Through comparative experiments, we find that the mapping network is the key to the output discrepancy and propose a novel initialization strategy of student, which can help resolve the output discrepancy issue.
The proposed latent-direction-based distillation loss further improves the distillation efficacy and we achieve state-of-the-art results in StyleGAN2/3 distillation, outperforming the rival method by a large margin on image realism, latent space smoothness and image projection fidelity.

\noindent
\textbf{Limitations.}
The computation and memory footprint of our method are larger than previous methods because it needs to compute the similarity between the original batch and transformed batch.
Besides,  we only consider the output discrepancy issue in unconditional GANs. 
In fact, this problem also exists in conditional setting when the condition is not strong enough (e.g., the conditional input is the class label).
How to analyze the output discrepancy issues of uncGANs and cGANs in a more general form is also a direction worth exploring.

% \noindent\textbf{Acknowledgement.}
% This work is supported by NTU NAP, MOE AcRF Tier 1 (2021-T1-001-088), and under the RIE2020 Industry Alignment Fund – Industry Collaboration Projects (IAF-ICP) Funding Initiative, as well as cash and in-kind contribution from the industry partner(s).

\noindent\textbf{Acknowledgement.}
This study is supported under the RIE2020 Industry Alignment Fund Industry Collaboration Projects (IAF-ICP) Funding Initiative, as well as cash and in-kind contribution from the industry partner(s). It is also supported by Singapore MOE AcRF Tier 2 (MOE-T2EP20120-0001). 

\clearpage
\bibliographystyle{splncs04}
\bibliography{egbib}

\newpage
\appendix
\setcounter{figure}{0}
\setcounter{table}{0}
\renewcommand\thefigure{\Alph{section}\arabic{figure}} 
\renewcommand\thetable{\Alph{section}\arabic{table}}

\section{Appendix}

\subsection{Implementation Details}

\textbf{Training Hyperparameters.}
For the mapping network mimicking of the first stage, we use Adam as the optimizer with a initial learning rate of 0.05.
We train for 50k steps and the batch size is set as 4096.
For the normal GAN training of the second stage, we use Adam optimizer with a initial learning rate of 0.002 and 450k iterations.
For the $\alpha$ that controls the offset along latent direction, we sample it from a Gaussian distribution $\mathcal{N}(0, 5)$.
We set $\lambda_{\mathrm{GAN}}$, $\lambda_{\mathrm{rgb}}$, $\lambda_{\mathrm{lpips}}$ and $\lambda_{\mathrm{LD}}$ to be 1, 3, 3 and 30, respectively.
The features that are used to compute LD loss come from the outputs of 64/128/256 resolution blocks.

\noindent
\textbf{Evaluation Metrics.}
Fr\'echet Inception Distance (FID) is a commonly used metric to evaluate the realism of generated images.
The generated images and real images are fed into a inception network and then a Fr\'echet distance is computed between their corresponding feature maps.
We use the implementation of FID in CAGAN~\cite{CAGAN}.
Specifically, we use 50K real images and 50K generated images to compute statistics, respectively.
Perceptual Path Length (PPL) is proposed in StyleGAN~\cite{stylegan} to measure the smoothness of latent space. We adopt the PPL implementation in CAGAN~\cite{CAGAN} for a fair comparison.
PSNR and LPIPS are used by CAGAN to evaluate the image projection ability.
A given real image is first mapped back to the latent space through optimizer such as L-BFGS.
The projected image is obtained by feeding this resulting latent code to the generator.
Then, the PSNR and LPIPS distance are computed between the projected image and the original image again.
A smaller value indicates that the generator can model the distribution in real world better.
We compute these two metrics using our own implementation.

\subsection{Distillation without GAN Loss}
\begin{figure*}[t]
	\centering
	\includegraphics[scale=0.36]{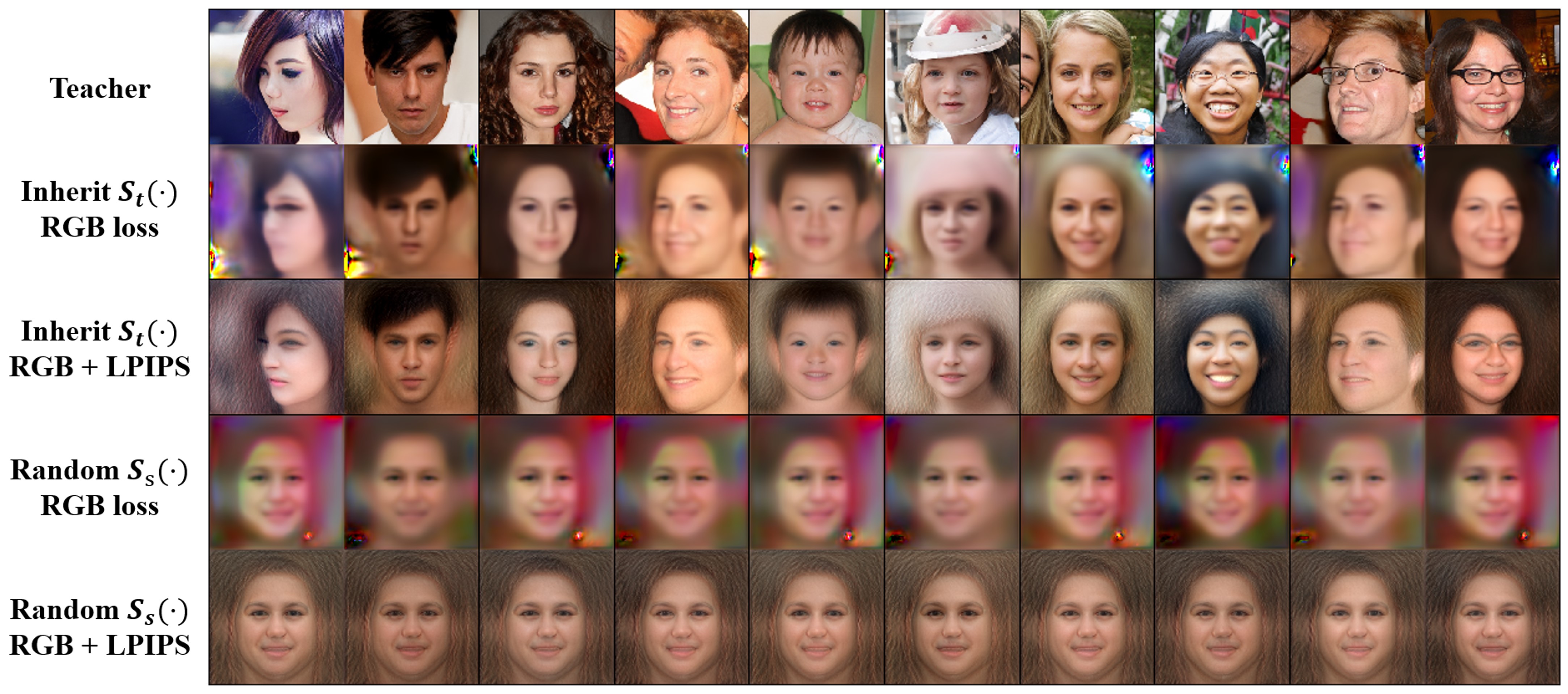}
	\vspace{-5pt}
	\caption{Distillation without GAN loss.}
	\label{fig:distill_super}
\end{figure*}

In Section. 3.3 of the main paper, we highlight that the mapping network decides whether a student can learn similar output to that of the teacher. 
To further examine this hypothesis, we train the student in a fully supervised manner.
Specifically, we remove the GAN loss and treat the $z$ and $G_t(z)$ as input/label pairs to train the student network.
The result is shown in Fig. \ref{fig:distill_super}.
It shows that the student cannot learn any meaningful content in the distillation process without a suitable mapping network. It yields the same face-like output for all the input noise.

\subsection{Latent-Direction-Based Distillation Loss}

\begin{table}[t]
    % \vspace{-5pt}
    \centering
    \caption{Ablation study about relation mimicking. Single View brings marginal improvement. Random Offset even has negative effect. Our LD loss consistently improves the performances of both RGB and RGB+LPIPS.}
    %\vspace{-5pt}
    \begin{tabular}{ccc}
        \toprule
        Mimicking Loss & $\mathcal{L}_{LD}$ & FID \\ 
        \midrule
        RGB  & N/A & 9.41 \\
        RGB + Random Offset & KL & 9.80 \\
        RGB + Single View  & KL & 9.47 \\
        RGB + LD  & L2 & 9.16 \\
        RGB + LD  & KL & 9.05 \\
        \midrule
        RGB + LPIPS &  N/A & 8.61 \\
        RGB + LPIPS + LD  & L2 & 8.64 \\
        RGB + LPIPS + LD  & KL & \textbf{8.26} \\
        \bottomrule
    \end{tabular}
    % \vspace{-5pt}
    \label{tab:ablation_LD}
\end{table}

The proposed latent-direction-based loss is essentially a relation loss.
We are interested in whether the benefit brought by $\mathcal{L}_{LD}$ comes from relation mimicking or from the latent-direction-based augmentation.
Specifically, we consider three variants: 1) Single View, namely the similarity is computed inside the normal samples rather than between normal samples and augmented samples, 2) Random Offset, namely we move $w$ along a random direction to get $f_i'$ instead of along the latent direction, 3) Our latent-direction-based method (abbreviated as LD).

\begin{figure}[t]
    \centering
    \subfloat[][In coarse style mixing, CAGAN generates glasses, which does not appear in both source images. CAGAN also produces blurry images in middle style case. In contrast, our style mixing results are more realistic and more similar to teacher.] {\includegraphics[scale=0.17]{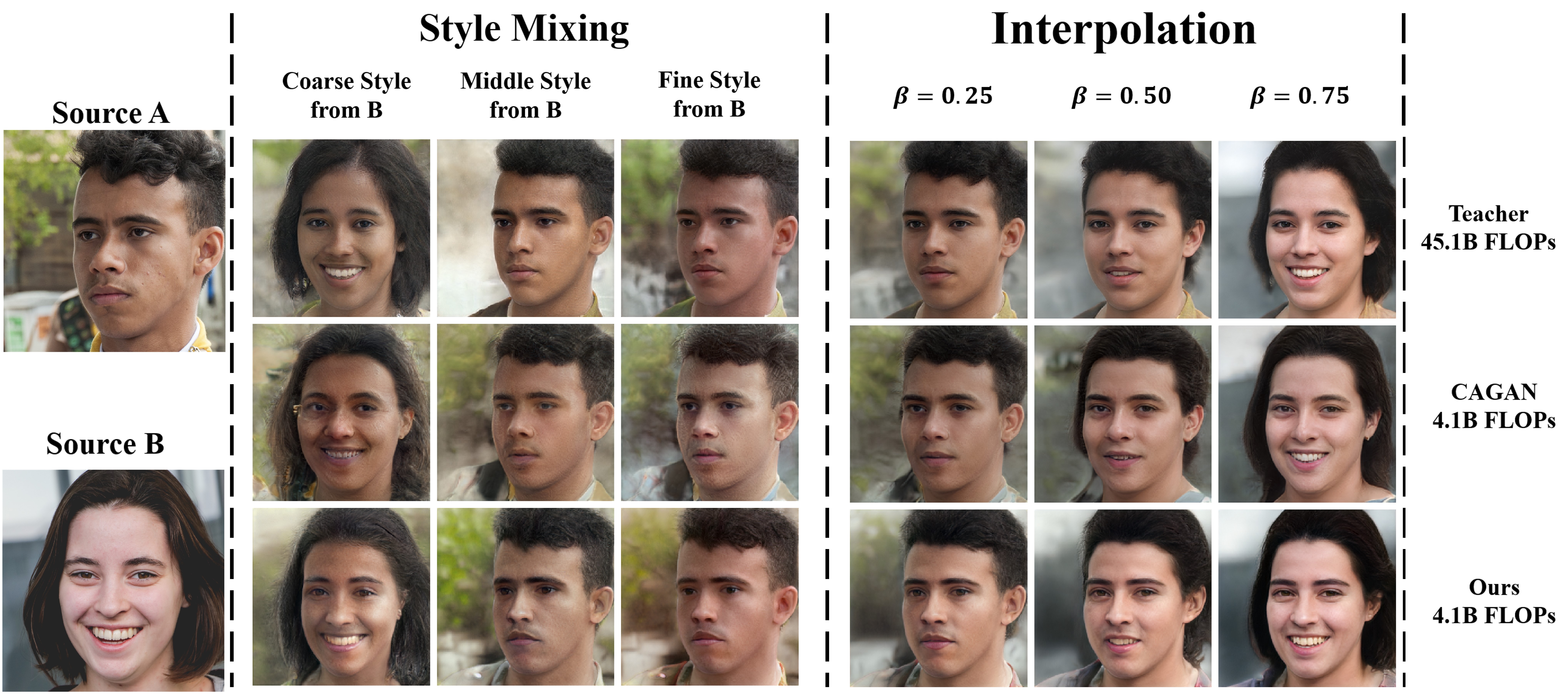}}\vfill\vspace{-10pt}
    \subfloat[][CAGAN generates lighting artifacts in coarse case and skin color artifacts in fine case, while our results are more realistic. In interpolation of CAGAN, the earrings disappear in $\beta=0.25$ but appear again in $\beta=0.50$. In contrast, our results are much smoother.] 
    {\includegraphics[scale=0.17]{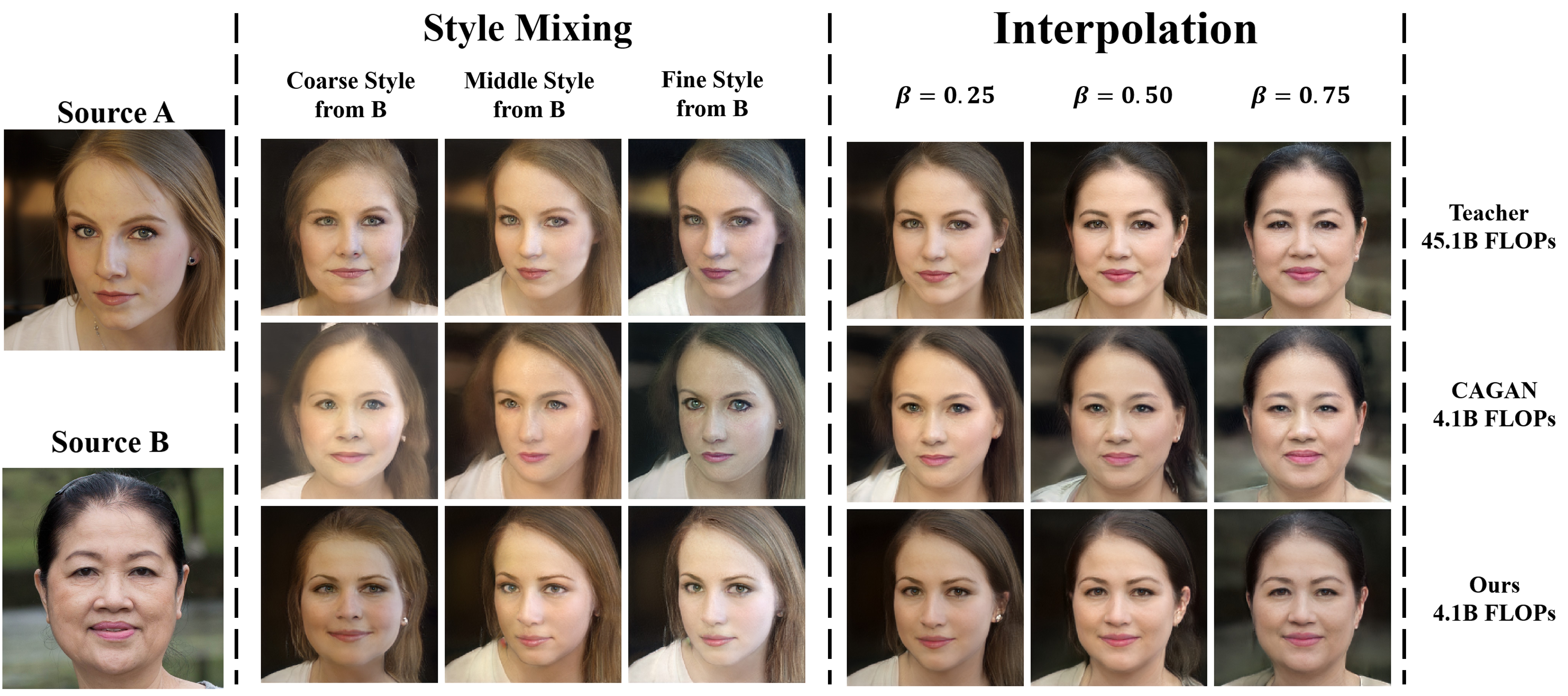}}\vfill
    \caption{Image editing results.}
    \label{fig:editing_supp}
\end{figure}

\subsection{Image Editing}
\label{sec:editing}

We demonstrate the superiority of our method on image editing, including style mixing and interpolation.
Given two real face images $I_A, I_B$, we first project them back to the latent space and get $w_A, w_B$. 
Both $w_A$ and $w_B$ are of shape $L\times D$, where $L$ is the number of convolution layers and $D$ is the dimension of latent code.
For style mixing, we replace the $i-$th vector in $w_A$ with that from $w_B$.
We set $i\in[1,3]$, $i\in[5,8]$ and $i\in[10,13]$ for coarse, middle and fine style mixing, respectively.
For interpolation, we linearly combine the latent code with $\beta$ controls the weight: $w = \beta\cdot w_A+(1-\beta)\cdot w_B$, and then feed $w$ into generator to get the interpolation results.
We edit the images on resolution 256$\times$256.

\subsection{StyleGAN2 Linear Separability}

\begin{figure}[t]
    \vspace{-15pt}
    \subfloat[][Original] {\includegraphics[scale=0.22]{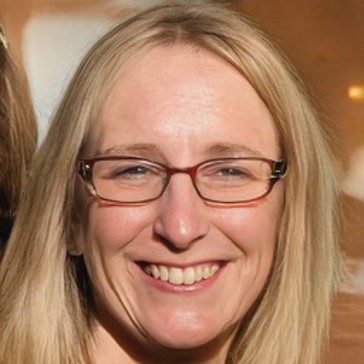}}\hfill
    \subfloat[][Single factor change] {\includegraphics[scale=0.22]{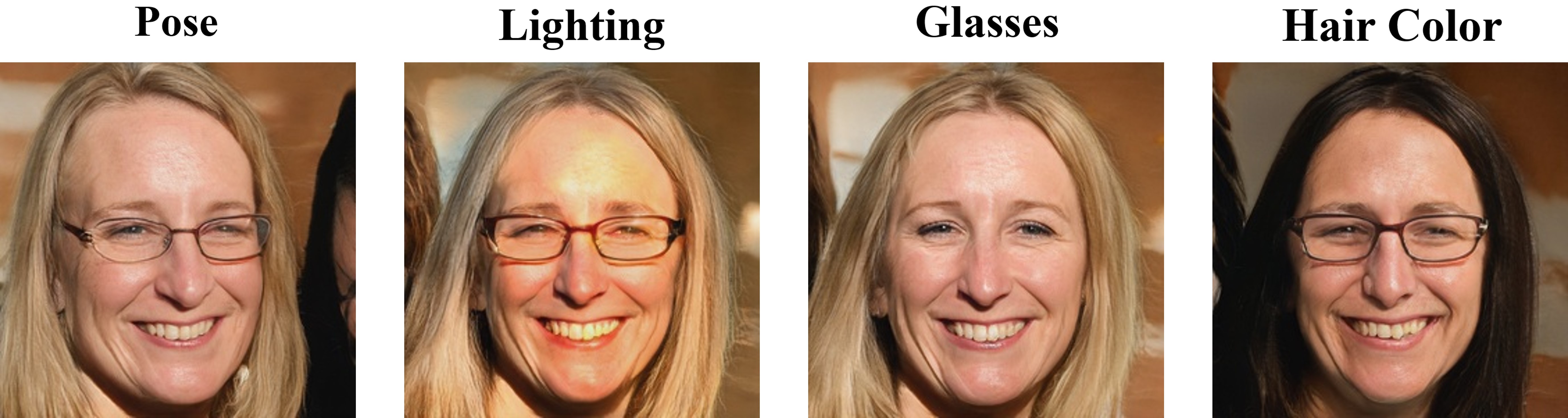}}\hfill
    \vspace{-5pt}
    \caption{StyleGAN2 shows good factorization in the $w$ space. It is possible to control a single semantic factor such as pose, lighting condition, glasses and hair color by moving the style vector $w$ of a certain layer along a specific direction.}
    \label{fig:sefa}
    \vspace{-5pt}
\end{figure}

A well-trained StyleGAN2 model is linear separable in the latent space. An example is shown in Fig.~\ref{fig:sefa}.
The results are shown in Fig.~\ref{fig:editing_supp}.
For style mixing, CAGAN always has artifacts in face shape (coarse style) and skin color (middle shape).
In contrast, the synthesized results of our method are more realistic and correspond better with two source images.
In the coarse style case, our result corresponds well on face shape and facial components with source B.
In the fine style case, our result corresponds well on lighting and skin color with source B.
For interpolation, we also observe a smoother change than CAGAN, showing that our method learns a better structure in the latent space.

\subsection{Image Projection}

\begin{figure*}[t]
	\centering
	\includegraphics[scale=0.07]{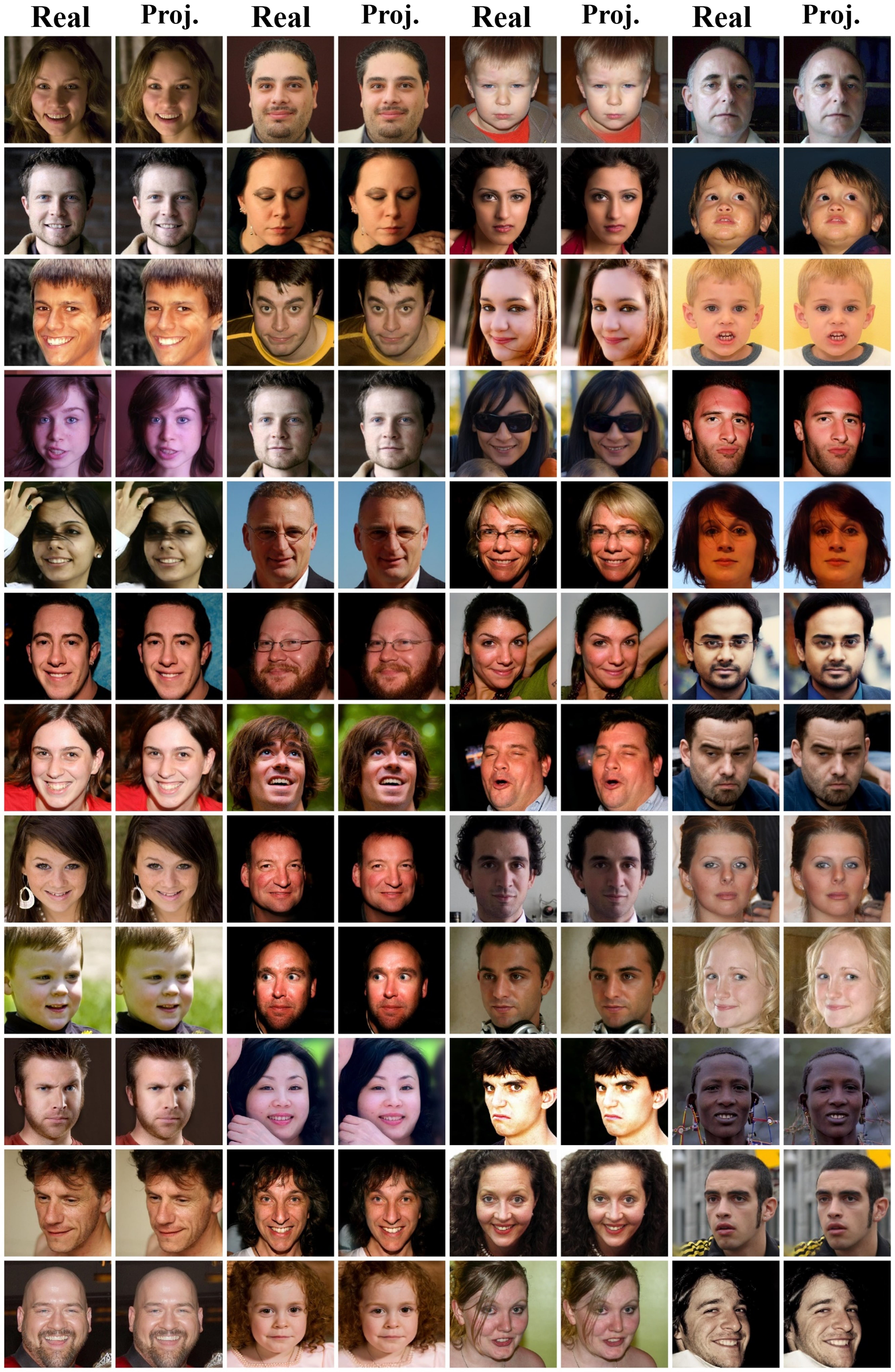}
	\vspace{-5pt}
	\caption{Image projection results. In each pair, the left image is from real world (not from training set) and the right image is the projected result by our model. Our method can model the real face distribution well.}
	\label{fig:projection}
\end{figure*}

We show image projection results of our method in Fig.~\ref{fig:projection}. All the real images come from Helen Set55~\cite{CAGAN} and are not seen in the training stage. Our model reconstructs them with high quality.

\subsection{Generation Results}\label{sec:qualitative}

\begin{figure*}[t]
	\centering
	\includegraphics[scale=0.04]{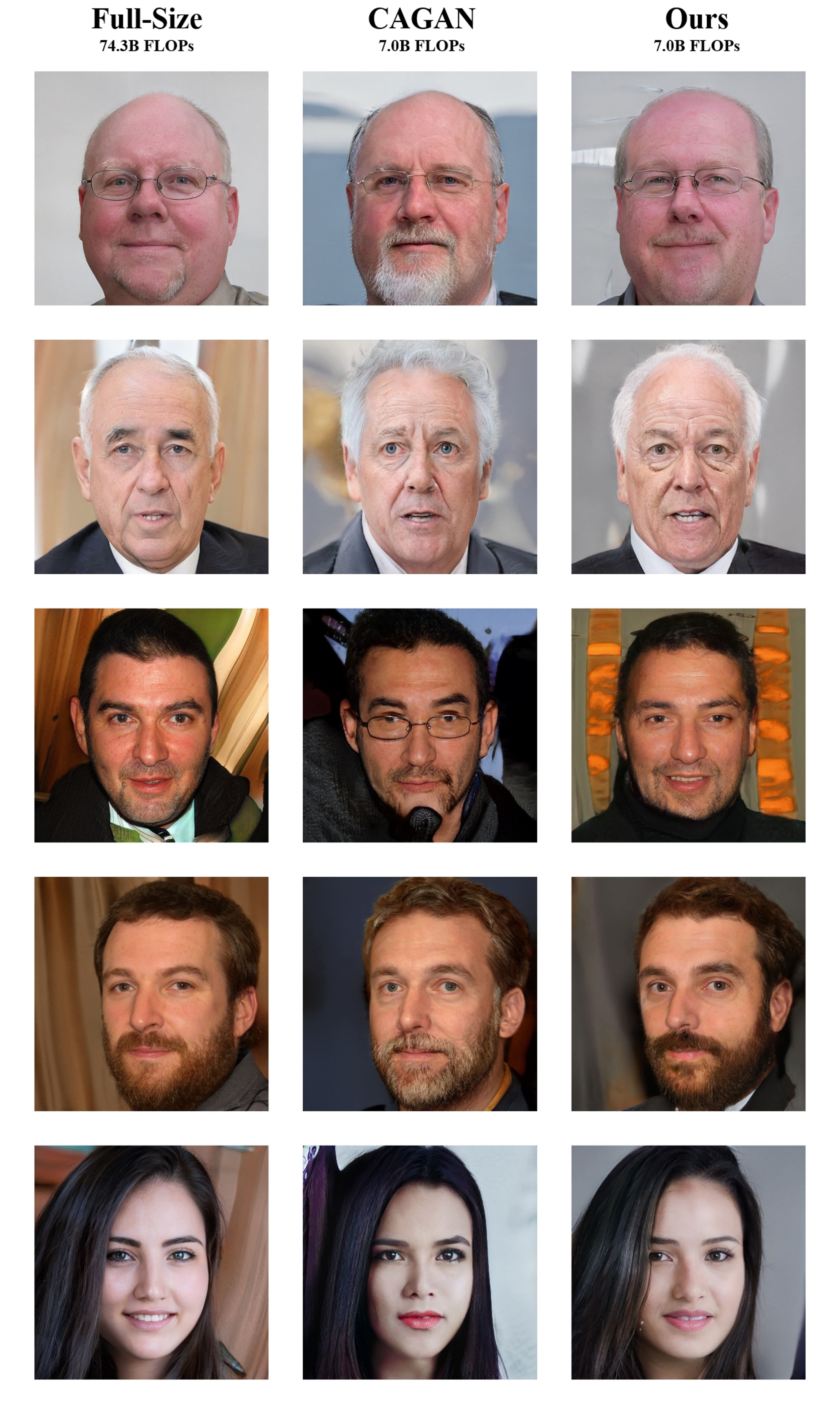}
	\vspace{-5pt}
	\caption{Generation results on resolution 1024$\times$1024. The synthesized images of Our method are of better quality than CAGAN. In several semantic factors such as beard, haircut and glasses, our results are more similar to the full-size model even though we do not inherit convolution weights.}
	\label{fig:compare_1024}
\end{figure*}

\begin{figure*}[t]
	\centering
	\includegraphics[scale=0.24]{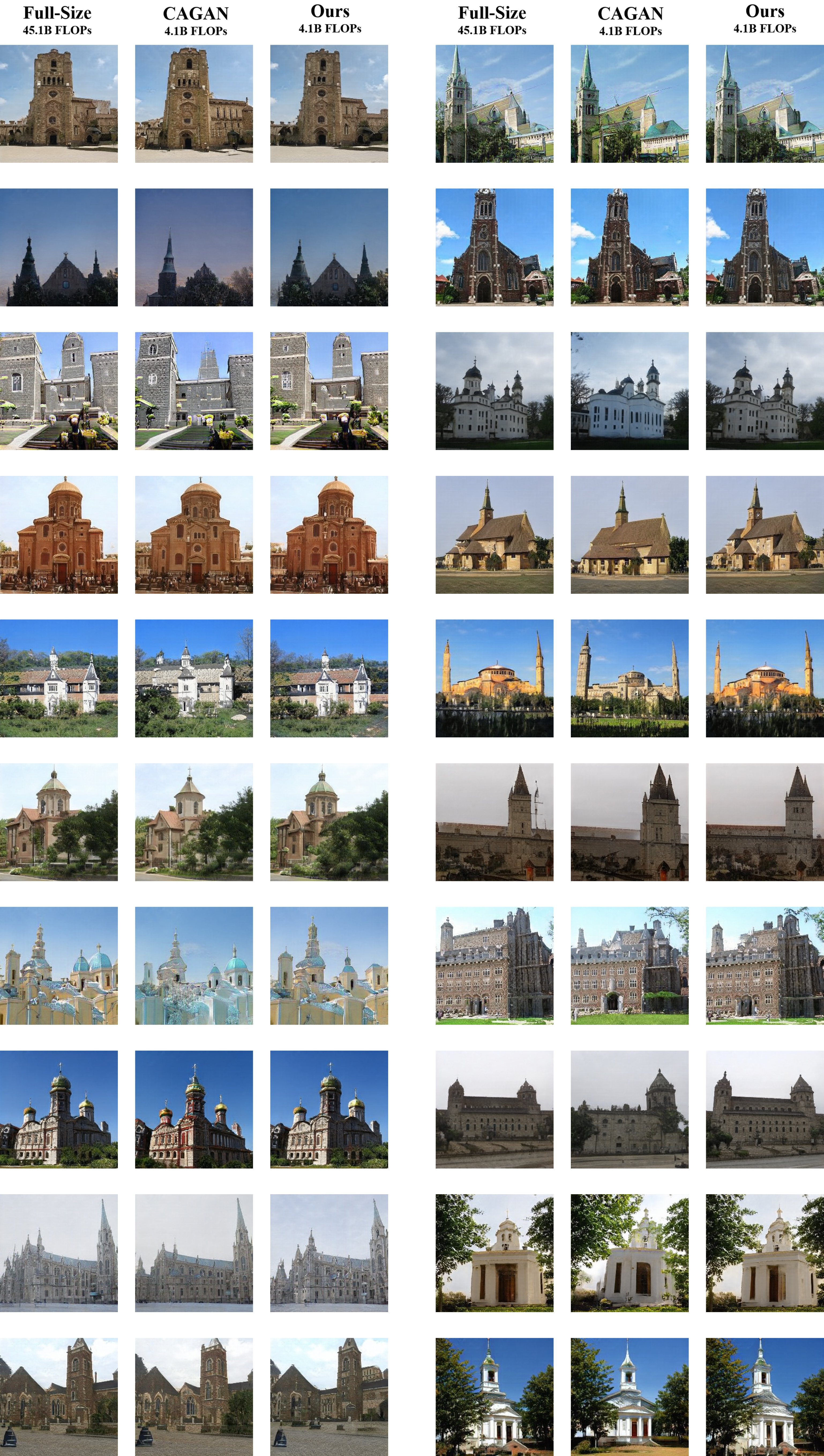}
	\vspace{-5pt}
	\caption{Generation results on LSUN church on resolution 256$\times$256.}
	\label{fig:compare_church}
\end{figure*}

We show more generation results of FFHQ and LSUN church datasets in Fig.~\ref{fig:compare_1024} and Fig.~\ref{fig:compare_church}, respectively.

% \begin{figure*}[t]
% 	\centering
% 	\includegraphics[scale=0.32]{compare_256_char.jpg}
% 	\vspace{-5pt}
% 	\caption{Synthesized results on resolution 256$\times$256. For Two-Stage, we compress the original 8-layer mapping network into 5 layers. The images of each row are generated using the same input noise $z$. Note that all the students are trained with mimicking loss. Random $S_s(z)$ cannotc generate images consistent with teacher since they have different mapping networks. Two-Stage mitigates the output discrepancy issue by directly mimicking the mapping network, but there are still semantic differences from teacher. Compared to CAGAN, our generated images have fewer artifacts and are more similar to teacher.}
% 	\label{fig:compare_256}
% \end{figure*}

\end{document}